\documentclass[final,5p,times,twocolumn,authoryear]{elsarticle}

\usepackage[utf8]{inputenc}
\usepackage[T1]{fontenc}
\usepackage[english]{babel}
\usepackage{amsmath,amssymb}
\usepackage{graphicx}
\usepackage{booktabs}
\usepackage{soul}
\usepackage{xcolor}
\usepackage{array}
\usepackage{makecell}
\usepackage[hidelinks]{hyperref}
\usepackage{dblfloatfix}
\usepackage{microtype}
\usepackage{balance}
\usepackage{prletter}

\journal{Pattern Recognition Letters}
\biboptions{authoryear,sort}

\begin{document}

\begin{frontmatter}

\title{Neoadjuvant chemotherapy response prediction using pretreatment diffusion and contrast-enhanced magnetic resonance imaging with clinical variables}

\tnotetext[prl]{This work is under consideration at Pattern Recognition Letters.}

\author[uniovi_cs,uniovi_bme]{Pablo García Marcos}
\ead{garciamarpablo@uniovi.es}
\author[uniovi_ee]{Paula Puerta González}
\ead{puertapaula@uniovi.es}
\author[unicoru_ma,unitx]{Guillermo Lorenzo}
\ead{guillermo.lorenzo@udc.es}
\author[purdue_me,purdue_bme,purdue_ccr]{Hector Gomez}
\ead{hectorgomez@purdue.edu}
\author[huca_ra]{Covadonga del Camino}
\ead{caminocovadonga@uniovi.es}
\author[cahu_onco]{Adán Rodríguez}
\ead{adan.rodriguez@sespa.es}
\author[cahu_onco]{Ignacio Peláez}
\ead{ignacio.pelaez@sespa.es}
\author[uniovi_cs,uniovi_bme]{Angel Rio-Alvarez\corref{cor1}}
\ead{rioangel@uniovi.es}
\author[uniovi_ee,uniovi_bme]{Víctor M. González}
\ead{vmsuarez@uniovi.es}
\cortext[cor1]{Corresponding author}

\address[uniovi_cs]{Computer Science Department, University of Oviedo, Campus de Viesques, 33203 Gij\'on, Asturias, Spain}
\address[uniovi_ee]{Electrical Engineering Department, University of Oviedo, Campus de Viesques, 33203 Gij\'on, Asturias, Spain}
\address[uniovi_bme]{Biomedical Engineering Center (BME), University of Oviedo, Campus de Viesques, 33203 Gij\'on, Asturias, Spain}
\address[unicoru_ma]{Group of Numerical Methods in Engineering, Department of Mathematics, University of A Coru\~na, Campus de Elvi\~na, 15071 A Coru\~na, Spain}
\address[unitx]{Oden Institute for Computational Engineering and Sciences, The University of Texas at Austin, 201 E 24th St, Austin, Texas 78712, USA}
\address[purdue_me]{School of Mechanical Engineering, Purdue University, 585 Purdue Mall, West Lafayette, Indiana 47907, USA}
\address[purdue_bme]{Weldon School of Biomedical Engineering, Purdue University, 206 S Martin Jischke Dr, West Lafayette, Indiana 47907, USA}
\address[purdue_ccr]{Purdue Institute for Cancer Research, Purdue University, 201 S University St, West Lafayette, Indiana 47907, USA}
\address[cahu_onco]{Oncology Service, Cabue\~nes University Hospital (CAHU), Calle Los Prados 395, 33394 Gij\'on, Asturias, Spain}
\address[huca_ra]{Radiodiagnostic Service, Asturias Central University Hospital (HUCA), Avenida de Roma s/n, 33011 Oviedo, Asturias, Spain}

\begin{abstract}
Prediction of pathological complete response before the start of neoadjuvant chemotherapy (NACT) may facilitate more tailored and effective therapeutic planning for breast cancer patients. This work proposes a deep-learning model designed exclusively for pretreatment data: apparent diffusion coefficient (ADC) maps, dynamic contrast-enhanced magnetic resonance imaging (DCE-MRI), and clinical variables. The study uses the public multicenter ACRIN 6698/I-SPY2 dataset for model training and evaluation. The architecture employs ImageNet-pretrained EfficientNet-B0 encoders for feature extraction and a simple classifier. Multiple clinical variables were evaluated: age, race, histological type, hormone receptor/human epidermal growth factor receptor 2 (HR/HER2) subtype, SBR grade, and maximum diameter. Only HR/HER2 subtype was retained because it was the only variable that consistently improved AUC. Using ten runs of five-fold cross-validation, standalone ADC and DCE-MRI achieved mean AUCs of 0.809 and 0.812, respectively. Adding HR/HER2 subtype increased the corresponding AUCs to 0.843 and 0.856. The final configuration, combining ADC, DCE-MRI, and HR/HER2 subtype, achieved the highest cross-validation AUC, 0.882. These results support the feasibility of baseline multimodal models for NACT response prediction, although the limited dataset size requires external validation before clinical use.
\end{abstract}
 
 
\end{frontmatter}
 
\section{Introduction}\label{sec:intro}
 
Breast cancer remains one of the main causes of oncological morbidity and mortality among women. For locally advanced or aggressive tumors, neoadjuvant chemotherapy (NACT) can reduce tumor volume, facilitate surgical intervention, and evaluate tissue sensitivity to systemic treatment \citep{Korde2021_ASCO_NACT_guideline,gianni_neoadjuvant_nodate,von_minckwitz_neoadjuvant_2012}. One of the main objectives of NACT is to achieve pathological complete response (pCR), defined as the absence of residual disease in the breast and lymph nodes. pCR is associated with better prognosis, especially in HER2-positive and triple-negative tumors \citep{cortazar_pathological_2014,von_minckwitz_pcr_definition_2012,esserman_pathologic_2012}. Nevertheless, not all patients present the desired response to NACT and, in the absence of reliable predictors, some are exposed to its toxicity without proportional benefit.

Response to NACT is also strongly influenced by tumor biology. HR/HER2 subtype captures clinically relevant differences in treatment
sensitivity and pCR rates \citep{von_minckwitz_pcr_definition_2012,
esserman_pathologic_2012}. Because this information is routinely available
before treatment, it may complement imaging biomarkers in baseline prediction
models.

Early prediction of pCR may be clinically relevant at two specific moments. During treatment, images acquired after one or several cycles may allow therapy to be adapted according to functional or volumetric changes \citep{Tudorica2016_EarlyPrediction_DCE,partridge_diffusion-weighted_2018}. Before treatment, baseline prediction may guide the initial intensity of the therapeutic strategy, select patients for adaptive therapies, or identify cases in which treatment should be intensified from the beginning \citep{liu_radiomics_2019,joo_multimodal_2021,wang_convolutional_2019}. Unlike longitudinal models, baseline prediction avoids waiting until the first cycle has been administered, but it is also significantly more challenging because it lacks dynamic information and requires tumor sensitivity to be inferred from pretreatment features.
 
Magnetic resonance imaging (MRI) has special relevance in breast cancer evaluation. Dynamic contrast-enhanced MRI (DCE-MRI) captures information about vascular perfusion and permeability and is therefore widely used in the diagnosis and monitoring of NACT \citep{hylton_neoadjuvant_2016,Tudorica2016_EarlyPrediction_DCE}. Diffusion-weighted MRI (DW-MRI) and its derived apparent diffusion coefficient (ADC) maps offer complementary characterization of water mobility and tumor cellularity and have shown predictive value \citep{partridge_diffusion-weighted_2018,Partridge2023_DWI_update}. While diffusion-only approaches favor simpler acquisition protocols without gadolinium contrast, multimodal models may achieve better performance by combining physiological information from DCE-MRI and diffusion with clinical variables.

In this study, we propose a baseline model of pCR using only information available before NACT.  Unlike previous work based on longitudinal DCE-MRI and DW-MRI/ADC information
\citep{Tudorica2016_EarlyPrediction_DCE,Liu2021_EarlyPrediction_SciRep,
partridge_diffusion-weighted_2018}, this work explores a multimodal pretreatment 
scenario integrating ADC maps, DCE-MRI, and clinical variables. Among the evaluated variables, HR/HER2 subtype was the only one that consistently improved
discrimination. ADC and DCE-MRI achieved comparable standalone AUCs (0.809 and 0.812), while adding HR/HER2 subtype increased the corresponding AUCs to 0.843 and 0.856. Combining ADC, DCE-MRI, and HR/HER2 subtype produced the best AUC, 0.882. Our main contributions are a standardized pretreatment imaging pipeline, a multimodal neural network integrating imaging
and clinical information, and an evaluation of the complementary value of each input source.

\section{Related work}\label{sec:related}
 
pCR prediction using MRI as the main input has traditionally been studied using radiomics and machine learning. These approaches extract texture, shape, and intensity features from DCE-MRI, DW-MRI/ADC, or other sequences to subsequently train supervised classifiers \citep{Tudorica2016_EarlyPrediction_DCE,liu_radiomics_2019}. Radiomics has shown that tumor heterogeneity contains predictive information, but it is sensitive to variations in acquisition protocols, scanners, and image processing, which may affect feature reproducibility and require appropriate standardization or harmonization strategies \citep{saint_martin_radiomics_2021}. Furthermore, several approaches integrate imaging features with clinicopathological variables, which can improve predictive performance but also increase the complexity and number of inputs of the predictive pipeline \citep{liu_radiomics_2019}.
 
Deep learning models reduce the need to manually design features. Convolutional networks trained using DCE-MRI or multimodal inputs have achieved competitive results for pCR prediction \citep{khan_deep_2022,Liu2021_EarlyPrediction_SciRep,joo_multimodal_2021}. Nevertheless, many of these techniques use images acquired during or after NACT, generating the prediction when part of the treatment has already been administered. Others combine multiple MRI sequences with clinical data \citep{joo_multimodal_2021}, although the resulting increase in input heterogeneity and model complexity may pose additional challenges for generalization across cohorts.
 
Studies of DW-MRI/ADC have shown that changes in diffusion during NACT are correlated with pCR \citep{partridge_diffusion-weighted_2018}. Multicenter studies such as ACRIN 6698 have supported the use of ADC as a biomarker, while recent reviews highlight the interest of contrast-free strategies to simplify protocols \citep{Mann2024_NonContrastBreastMRI,Partridge2023_DWI_update}. Still, in a strictly baseline scenario, ADC alone may fail to capture vascular or perfusion features present in DCE-MRI. Thus, a middle ground should be pursued, using baseline models that combine only ADC and DCE-MRI with limited clinical data to provide sufficient but controlled information.
 
\section{Materials and methods}\label{sec:methods}
 
\subsection{Dataset}\label{subsec:dataset}
 
The public ACRIN 6698/I-SPY2 dataset from The Cancer Imaging Archive was used \citep{noauthor_acrin_nodate}. The ACRIN 6698 trial contains MRI studies of breast cancer patients treated with NACT, featuring acquisitions at several time points: before treatment (T0), after three weeks of paclitaxel (T1), between paclitaxel and anthracycline therapy (T2), and at the end of treatment (T3). Each session includes DCE-MRI, DW-MRI, ADC maps, and expert-labeled tumor segmentations for diffusion/ADC, together with acquisition details and clinical variables \citep{newitt_testretest_2019,partridge_diffusion-weighted_2018}. In this study, only T0 was used, as the objective was to predict pCR before patient exposure to treatment. The images used as DCE-MRI input belong to the third post-contrast acquisition, which was selected a priori because derived tumor segmentation masks were available for this phase and it provides sufficient enhancement, rather than on the basis of model performance. This selection was not evaluated during validation, but instead was fixed before training to ensure simple and comparable baseline representation between patients. Nevertheless, given that the selected temporal phase in DCE-MRI may affect the information extracted by the model, this decision is considered a methodological simplification and a potential limitation. The DCE-MRI tumor masks used in this study were obtained from a Zenodo resource containing post-processed tumor segmentation masks derived from the ACRIN 6698 collection and represented on the native image grid \citep{corral_fontecha_2026}. 
 
After applying availability and consistency criteria between modalities, a final cohort of 134 patients was used, including 89 non-pCR patients and 45 pCR patients. The pCR label was encoded as the positive class. The ADC and DCE-MRI images were organized by patient and class, keeping only those with available clinical variables and segmentation masks for both ADC and DCE-MRI.
 
\subsection{Image preprocessing}\label{subsec:preprocessing}

\begin{figure}[t]
\centering
\includegraphics[width=0.80\linewidth]{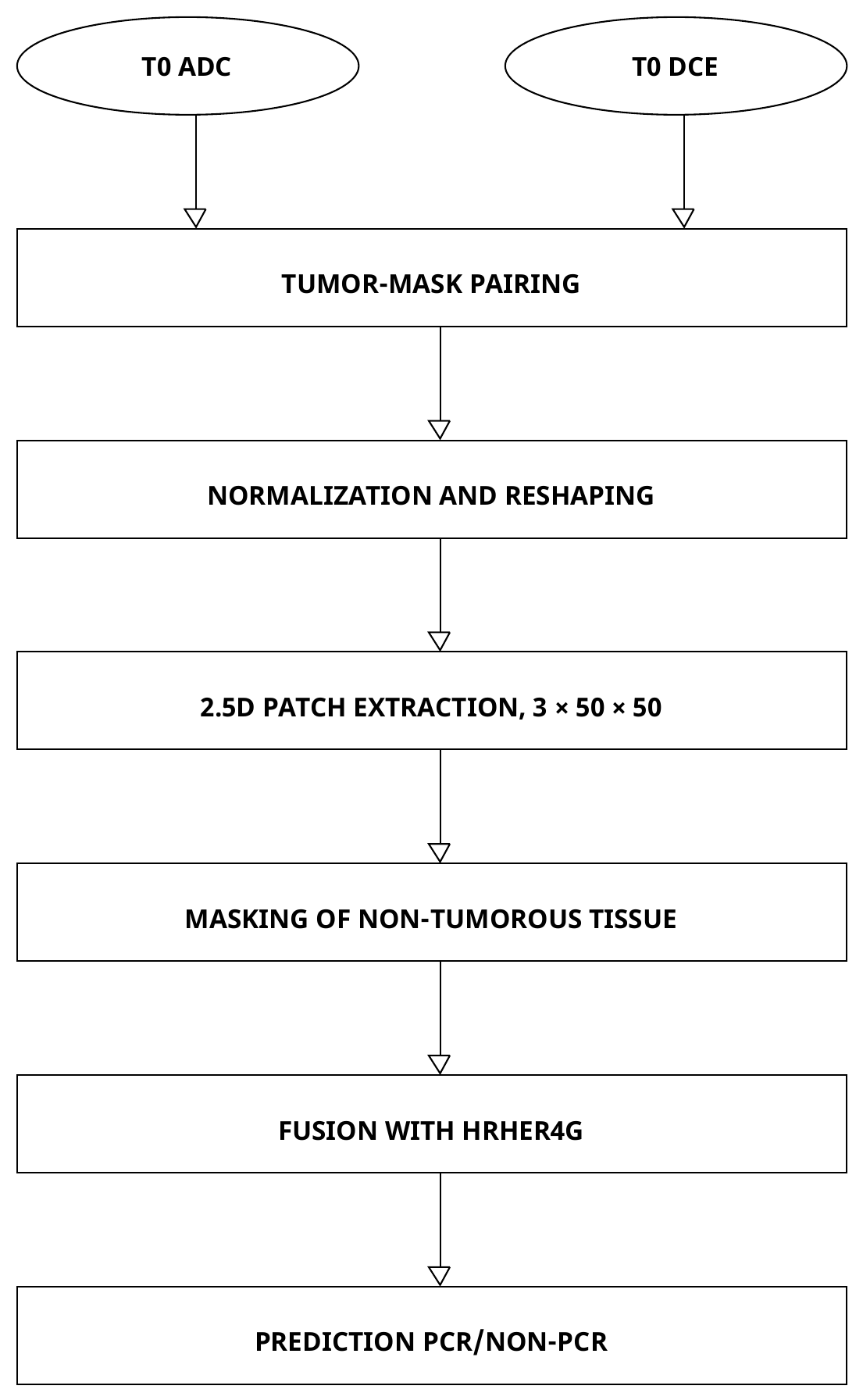}
\caption{Baseline preprocessing and multimodal fusion. All modalities are restricted to a comparable tumor patch; HRHER4G denotes the encoded HR/HER2 subtype.}
\label{fig:pipeline}
\end{figure}

Fig.~\ref{fig:pipeline} summarizes the baseline imaging preprocessing workflow and multimodal fusion. First, the images were paired with the corresponding tumor masks using patient identifiers. The images were resized and normalized through intensity scaling. Then, volumes were reconstructed for each patient and modality, and a patch centered on the tumor region was extracted. The input was built as a 2.5D representation by selecting three consecutive slices centered on the area with the largest tumor burden according to the segmentation mask. These three slices were stacked as input channels, creating patches of size $3\times50\times50$.
 
This strategy preserves limited inter-slice anatomical context while remaining compatible with convolutional encoders pretrained on RGB images. Furthermore, the use of three slices for each patient and modality reduces the high computational cost and overfitting risk associated with fully three-dimensional networks \citep{zhang_bridging_2022}. The $50\times50$ pixel patch size was used as a compromise between preserving tumor tissue and limiting the presence of irrelevant context.
 
Subsequently, non-tumor tissue was removed from the images by setting to zero all non-tumor pixels except for a margin of 5 pixels around the lesion. This margin was fixed to 5 pixels after visual inspection as a compromise between maintaining peritumoral information and limiting the presence of healthy background tissue. The margin was fixed before model training and was not optimized using validation performance. This process was performed independently for both ADC and DCE.
 
\subsection{Clinical variables}\label{subsec:clinical}
 
Multiple clinical variables were analyzed: age (\texttt{age}), race 
(\texttt{race}), lesion type (\texttt{Ltype}), hormone receptor/human 
epidermal growth factor receptor 2 (HR/HER2) subtype (\texttt{hrher4g}), 
Scarff--Bloom--Richardson (SBR) grade (\texttt{SBRgrade}), and 
MRI longest diameter (MRLD; \texttt{MRLD}). Numerical variables (\texttt{age}, \texttt{MRLD}) were standardized within each training partition. Categorical variables (\texttt{race}, \texttt{Ltype}, HR/HER2 subtype, \texttt{SBRgrade}) were encoded using one-hot encoding, with fixed global categories to avoid shape changes between folds. Multiple variable combinations were empirically tested, with HR/HER2 subtype being the only one that achieved consistent improvement in the area under the receiver operating characteristic curve (AUC) during preliminary testing. Thus, in the final results, ``clinical variables'' refers specifically to HR/HER2 subtype.
 
\subsection{Multimodal architecture}\label{subsec:architecture}
 
\begin{figure}[t]
\centering
\includegraphics[width=0.80\linewidth]{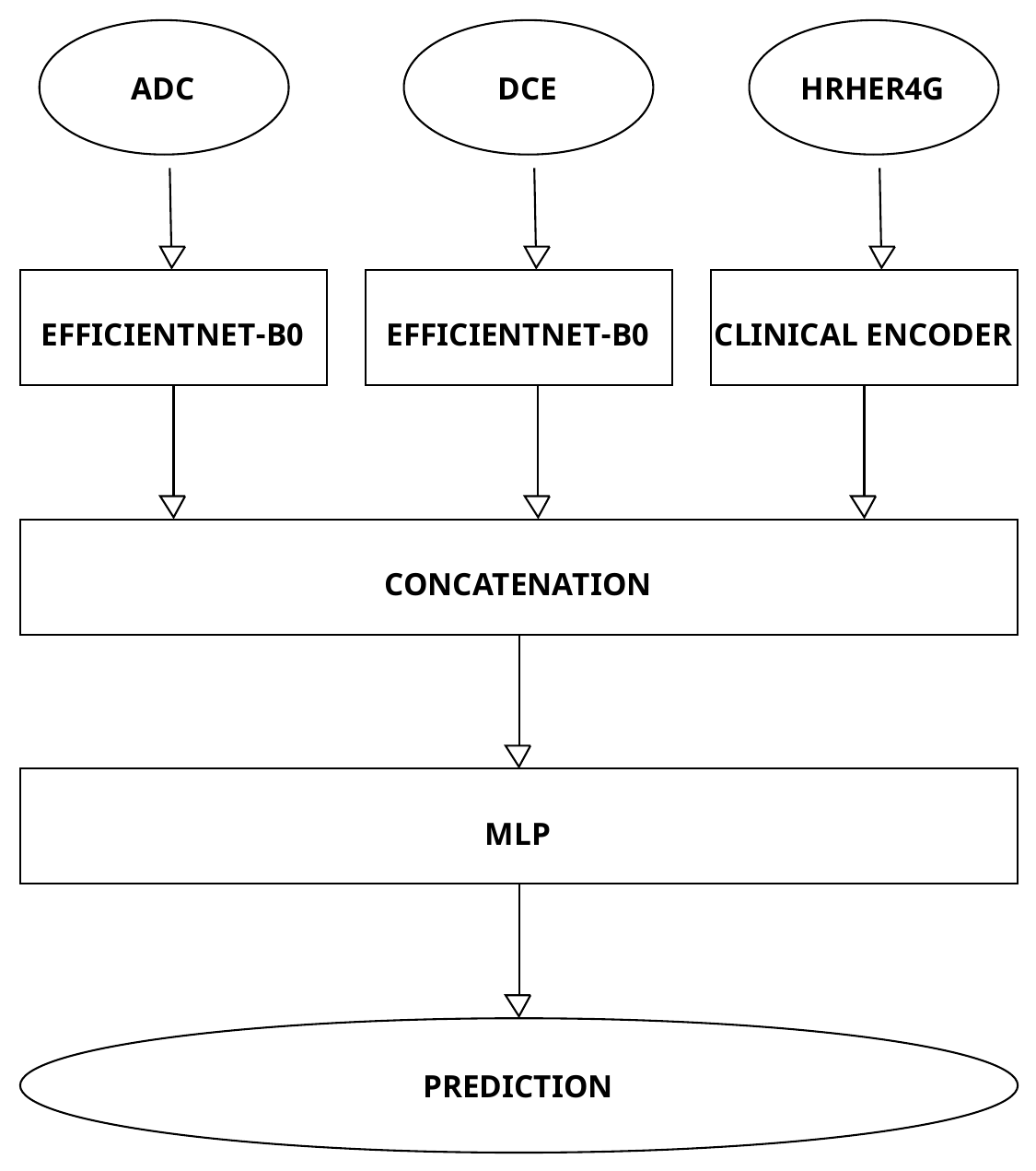}
\caption{Fusion architecture used by the multimodal configurations; HRHER4G denotes the encoded HR/HER2 subtype.}
\label{fig:fusion}
\end{figure}

The proposed network employs late feature fusion, as shown in Fig.~\ref{fig:fusion}. Each imaging modality is processed by an independent EfficientNet-B0 encoder initialized with ImageNet-pretrained weights. The resulting feature vectors are normalized using LayerNorm before the ADC and DCE-MRI vectors are concatenated. Clinical variables are processed using a linear layer followed by ReLU activation, dropout, and normalization. The clinical and imaging vectors are then combined in a classifier comprising two linear layers with ReLU and a dropout rate of 0.4 to produce the pCR/non-pCR probabilities.
 
\subsection{Training and metrics}\label{subsec:training}
 
The experiments were carried out through ten runs of stratified 5-fold cross-validation with seeds 42--51. In each fold, patients were split into training and validation subsets, using 20\% for  validation. Stratification was performed to ensure a similar proportion of pCR and non-pCR patients in each partition. Given the limited cohort size, especially regarding the pCR patients, each validation split contained few positive instances. Thus, several strategies were used to reduce overfitting and avoid data leakage.

The standardization parameters for numerical variables and the encodings for categorical variables were fitted exclusively on the training data of each fold and then applied to the corresponding validation subset.

Data augmentation was applied on the fly only to the training subset of each fold, without applying random transformations to the validation subsets. The augmentations included horizontal and vertical flips, rotations of up to $\pm10^\circ$, scaling variations, translations, and small adjustments to brightness and contrast.

Models were trained for up to 100 epochs using the AdamW optimizer with an initial learning rate of $10^{-4}$ and a weight decay of $10^{-4}$. A weighted cross-entropy loss function was used during training.

The strategies used to reduce overfitting included EfficientNet-B0 encoders pretrained on ImageNet, dropout in the clinical-variable encoder and final classifier, data augmentation, early stopping, and learning-rate reductions when validation performance plateaued.

Seven input configurations were evaluated: ADC, DCE-MRI, clinical variables, ADC+DCE-MRI, ADC+clinical variables, DCE-MRI+clinical variables, and ADC+DCE-MRI+clinical variables. Performance was evaluated using accuracy, weighted F1 score, AUC, sensitivity, specificity, positive predictive value (PPV), and negative predictive value (NPV). For each metric, the five validation-fold values were averaged within each run; the reported mean and standard deviation were then calculated across the ten run means. Threshold-based metrics used the class with the highest predicted probability, and undefined ratios were excluded from the averages. The reported metrics are internal validation results; no independent external validation cohort was available.
 
\section{Results}\label{sec:results}

Table~\ref{tab:results} displays the mean validation performance and standard deviation across ten runs of the evaluated configurations. As standalone imaging inputs, ADC and DCE-MRI achieved similar AUCs (0.809 and 0.812, respectively). ADC yielded a higher F1 score (0.682 vs.\ 0.660) and accuracy (72.7\% vs.\ 69.8\%). The clinical-only configuration obtained the lowest values for these three metrics.
 
\begin{table*}[t]
\centering
\caption{Validation performance across ten runs of five-fold cross-validation. Values are mean $\pm$ SD across run means. Acc, Sens, Spec, PPV, and NPV are percentages; F1 is class-frequency weighted. ``Clinical variables'' denotes the HR/HER2 subtype selected in the preliminary analysis. The best mean in each metric column is shown in bold.}
\label{tab:results}
\small
\setlength{\tabcolsep}{3pt}
\resizebox{\textwidth}{!}{%
\begin{tabular}{lccccccc}
\toprule
\textbf{Input} & \textbf{AUC} & \textbf{Acc (\%)} & \textbf{F1} & \textbf{Sens (\%)} & \textbf{Spec (\%)} & \textbf{PPV (\%)} & \textbf{NPV (\%)} \\
\midrule
ADC & $0.809 \pm 0.037$ & $72.7 \pm 2.8$ & $0.682 \pm 0.034$ & $33.4 \pm 8.9$ & $91.1 \pm 2.7$ & $73.1 \pm 6.1$ & $75.4 \pm 2.9$ \\
DCE-MRI & $0.812 \pm 0.036$ & $69.8 \pm 3.2$ & $0.660 \pm 0.044$ & $43.1 \pm 13.7$ & $82.3 \pm 7.6$ & $62.7 \pm 8.8$ & $77.4 \pm 4.3$ \\
Clinical variables & $0.736 \pm 0.049$ & $66.6 \pm 2.8$ & $0.594 \pm 0.034$ & $23.1 \pm 9.9$ & $86.9 \pm 6.6$ & $54.4 \pm 13.3$ & $71.8 \pm 2.9$ \\
ADC + DCE-MRI & $0.814 \pm 0.027$ & $74.3 \pm 2.3$ & $0.718 \pm 0.028$ & $\mathbf{46.6 \pm 9.4}$ & $87.2 \pm 4.2$ & $65.5 \pm 8.8$ & $78.8 \pm 2.8$ \\
ADC + clinical variables & $0.843 \pm 0.029$ & $74.3 \pm 3.3$ & $0.691 \pm 0.048$ & $31.7 \pm 10.4$ & $94.1 \pm 3.2$ & $77.2 \pm 9.5$ & $75.7 \pm 3.1$ \\
DCE-MRI + clinical variables & $0.856 \pm 0.027$ & $73.4 \pm 3.7$ & $0.676 \pm 0.054$ & $28.0 \pm 11.0$ & $\mathbf{94.5 \pm 4.6}$ & $70.9 \pm 18.0$ & $74.5 \pm 2.7$ \\
ADC + DCE-MRI + clinical variables & $\mathbf{0.882 \pm 0.020}$ & $\mathbf{77.3 \pm 1.6}$ & $\mathbf{0.741 \pm 0.024}$ & $44.9 \pm 7.8$ & $92.4 \pm 2.9$ & $\mathbf{77.8 \pm 5.2}$ & $\mathbf{79.3 \pm 2.3}$ \\
\bottomrule
\end{tabular}}
\end{table*}
 
Combining the two imaging modalities increased the mean AUC to 0.814 and the mean F1 score to 0.718. Adding HR/HER2 subtype increased the mean AUC from 0.809 to 0.843 for ADC and from 0.812 to 0.856 for DCE-MRI. The remaining clinical variables, including age, race, histological type, SBR grade, and maximum diameter, did not produce an average AUC improvement during the preliminary experiments.

The final ADC+DCE-MRI+HR/HER2 subtype configuration achieved the highest mean AUC, $0.882 \pm 0.020$, together with the highest accuracy, $77.3 \pm 1.6$\%, and F1 score, $0.741 \pm 0.024$. It also achieved a sensitivity of $44.9 \pm 7.8$\%, specificity of $92.4 \pm 2.9$\%, PPV of $77.8 \pm 5.2$\%, and NPV of $79.3 \pm 2.3$\%. The highest sensitivity was obtained with ADC+DCE-MRI ($46.6 \pm 9.4$\%), whereas DCE-MRI+HR/HER2 subtype produced the highest specificity ($94.5 \pm 4.6$\%). Thus, the all-source configuration provided the best ranking-based discrimination according to AUC, although it did not maximize every threshold-dependent metric.

\section{Discussion}\label{sec:discussion}

ADC and DCE-MRI provided comparable baseline discrimination when used independently. DCE-MRI achieved a slightly higher AUC, whereas ADC achieved higher F1 and accuracy values. Combining both modalities increased the AUC and F1 score relative to either standalone input, suggesting that enhancement and perfusion-related information is complementary to the cellularity and water-diffusion information represented by ADC maps.

Although the clinical-only model showed limited performance, incorporating HR/HER2 subtype increased AUC for ADC, DCE-MRI, and their combination. This finding is consistent with the biological relevance of receptor status in the response to NACT \citep{von_minckwitz_pcr_definition_2012,esserman_pathologic_2012}. Adding HR/HER2 subtype increased accuracy and F1 for each imaging configuration and the all-source configuration achieved the highest values for both metrics. The remaining clinical variables did not improve mean AUC.

The incorporation of a second image encoder increases the number of trainable parameters. With a cohort of 134 patients, small differences between cross-validation partitions may therefore affect the resulting AUC. The standard deviation observed across runs and the limited number of positive cases in each validation partition reinforce the need to interpret the results as internal validation rather than as an estimate of performance in an independent clinical population.

The comparison with previous work must account for the time point used to generate the prediction. Models using post-baseline or longitudinal information have access to treatment-induced changes, which provides a predictive advantage over strictly pretreatment approaches \citep{krasniqi_multimodal_2025}. These studies should therefore be considered reference points rather than direct comparisons with the proposed baseline model. Table~\ref{tab:comparison_longitudinal} summarizes studies based on ACRIN 6698, BMMR2, or related longitudinal settings.

Within ACRIN 6698, conventional ADC analysis showed moderate discriminative capacity using mid-treatment or post-treatment information, with reported AUC values of 0.60 and 0.61, respectively \citep{partridge_diffusion-weighted_2018}. The BMMR2 challenge reported AUC values of 0.803--0.840 using DCE-MRI and/or DW-MRI information \citep{li_breast_2024}. More recent longitudinal approaches, including the MRI-based enhanced self-attention network combined with clinicopathological characteristics and physiologically decomposed diffusion-weighted imaging, achieved AUC values close to 0.89--0.90 \citep{du_early_2025,gilad_radiomics_2025}. However, these approaches use images acquired after treatment initiation. In contrast, the proposed model achieved a mean validation AUC of 0.882 using exclusively pretreatment data.

\begin{table*}[t]
\centering
\caption{Comparison with MRI-based studies using post-baseline or longitudinal information for pCR prediction.}
\label{tab:comparison_longitudinal}
\footnotesize
\setlength{\tabcolsep}{2.5pt}
\renewcommand{\arraystretch}{1.0}
\begin{tabular}{p{0.12\textwidth}p{0.15\textwidth}p{0.18\textwidth}p{0.16\textwidth}p{0.10\textwidth}p{0.20\textwidth}}
\toprule
\textbf{Study} &
\textbf{Dataset} &
\textbf{Input data} &
\textbf{Prediction time point} &
\textbf{AUC} &
\textbf{Main comment} \\
\midrule
\citet{partridge_diffusion-weighted_2018} &
ACRIN 6698 &
DW-MRI/ADC &
Mid-/post-treatment &
0.60 / 0.61 &
Monitoring biomarker; not baseline-only prediction \\
 
\citet{li_breast_2024} &
BMMR2 / ACRIN 6698-derived &
DCE-MRI and/or DW-MRI, optional clinical variables &
Mostly post-baseline &
0.803--0.840 &
Challenge setting using information after treatment onset \\
 
\citet{du_early_2025} &
ACRIN 6698 + external cohort &
T1-DCE + DW + ADC + clinicopathological variables &
Pre- and early-treatment longitudinal MRI &
0.903 / 0.861 &
Not baseline-only; includes external testing \\
 
\citet{gilad_radiomics_2025} &
BMMR2 / ACRIN 6698-derived &
Physiologically decomposed DW-MRI radiomics &
Longitudinal early-/mid-treatment MRI &
0.89 &
Early- and mid-treatment scans are used \\
 
Current study &
ACRIN 6698 &
ADC + DCE-MRI + HR/HER2 subtype &
Pretreatment &
0.882 &
Strictly baseline prediction \\
\bottomrule
\end{tabular}
\end{table*}

Among strictly pretreatment studies, the comparison is more direct, although the reported values arise from different validation designs. \citet{liu_radiomics_2019} reported an AUC of 0.86 in their development cohort using multiparametric MRI radiomics and clinical variables; the AUCs in three independent external-center cohorts ranged from 0.71 to 0.80. The proposed model achieved a mean internal-validation AUC of 0.882 across ten runs using ADC, DCE-MRI, and HR/HER2 subtype, without T2-weighted MRI or handcrafted radiomic feature extraction, although these values should not be interpreted as a direct comparison.

\citet{joo_multimodal_2021} reported an AUC of 0.888 on an internal 80/20 holdout using pretreatment T1-weighted subtraction images, T2-weighted images, and clinical variables in a single-center cohort of 536 patients. This cohort was substantially larger than the cohort used in the present study, but no independent external test cohort was used. \citet{wang_convolutional_2019} reported, in a convolutional restricted Boltzmann machine (CRBM)-based radiomics preprint, a pretreatment AUC of 0.87 for the HER2-positive/triple-negative subset using DCE-MRI and internal four-fold cross-validation. That paper does not clearly document patient-level separation of the image slices across folds. Table~\ref{tab:comparison_baseline} summarizes these pretreatment studies.

\begin{table*}[t]
\centering
\caption{Comparison with MRI-based studies using strictly pretreatment information for pCR prediction. Development and validation results are distinguished where reported.}
\label{tab:comparison_baseline}
\footnotesize
\setlength{\tabcolsep}{3pt}
\renewcommand{\arraystretch}{1.0}
\begin{tabular}{p{0.13\textwidth}p{0.16\textwidth}p{0.22\textwidth}p{0.07\textwidth}p{0.10\textwidth}p{0.22\textwidth}}
\toprule
\textbf{Study} &
\textbf{Dataset} &
\textbf{Input data} &
\textbf{AUC} &
\textbf{Validation} &
\textbf{Main comment} \\
\midrule
\citet{liu_radiomics_2019} &
Multicenter non-ACRIN cohort &
DCE + DWI + T2 radiomics + clinical variables &
0.86 development; 0.71--0.80 external &
External, three site-held-out cohorts &
Handcrafted radiomics; independent-center validation \\
 
\citet{joo_multimodal_2021} &
Single-center non-ACRIN cohort (536 patients) &
T1-weighted subtraction MRI + T2-weighted MRI + clinical variables &
0.888 internal holdout &
Internal random 80/20 holdout &
Large single-center cohort; no independent test cohort \\
 
\citet{wang_convolutional_2019} &
Small non-ACRIN cohort &
DCE-MRI CRBM-radiomics &
0.87 internal CV &
Internal 4-fold CV; no external test &
Preprint; patient-level slice separation not documented \\
 
Current study &
ACRIN 6698 (134 patients) &
ADC + DCE-MRI + HR/HER2 subtype &
$0.882 \pm 0.020$ internal validation &
Ten runs of patient-level 5-fold CV; validation metrics; pooled sites &
Public multicenter dataset; no site-held-out test \\
\bottomrule
\end{tabular}
\end{table*}

Several limitations should be considered. First, the cohort remains limited for training a multimodal deep learning model with two EfficientNet-B0 encoders. Although stratified cross-validation, data augmentation, dropout, class weighting, early stopping, and fold-specific preprocessing were used, these strategies do not replace validation in an external cohort.

Second, the method depends on tumor masks reconstructed from the available annotations. Its clinical implementation would require a robust automatic or semi-automatic segmentation procedure and an evaluation of the sensitivity of the predictions to segmentation variability.

Finally, DCE-MRI requires administration of a gadolinium-based contrast agent. Its contribution to prediction performance must therefore be considered in relation to acquisition time, cost, and contrast-related restrictions. In settings where DCE-MRI is routinely acquired, the all-source model may provide improved baseline discrimination. In contrast-free workflows, ADC combined with HR/HER2 subtype may represent a simpler alternative, with an AUC of 0.843 compared with 0.882 for the all-source configuration.

\section{Conclusions}\label{sec:conclusions}
 
This work presents a baseline multimodal model capable of predicting NACT response in breast cancer using ADC, DCE-MRI, and the clinical variable HR/HER2 subtype. ADC and DCE-MRI achieved comparable standalone performance, and combining the two imaging modalities improved AUC and F1. Incorporating HR/HER2 subtype increased AUC for each imaging configuration. The combined ADC+DCE-MRI+HR/HER2 subtype configuration achieved the highest mean validation AUC, 0.882, and the highest accuracy and F1 point estimates across ten runs. These results display the potential of deep learning with pretreatment tumor patches to estimate response before the beginning of NACT.
 
Future work should focus on external validation and interpretability. Using external cohorts would help validate the model performance for clinical implementation. Further testing of image preprocessing could also be useful to determine which patch regions and modalities extract the most relevant features for prediction. Work on semantic segmentation could also contribute to this research, as reliable automatic semantic segmentations would facilitate real-world implementation of the proposed technique.
 
\section*{Acknowledgments}
 
This research was funded in part by the Council of Gijón through the University Institute of Industrial Technology of Asturias (IUTA) grants SV-25-GIJON-1-14, SV-25-GIJON-1-02, SV-24-GIJON-1-05, SV-24-GIJON-1-18, SV-24-GIJON-1-16, SV-23-GIJON-1-09, SV-22-GIJON-1-19, and SV-21-GIJON-1-19, and by Principado de Asturias, grant SV-PA-21-AYUD/2021/50994. GL acknowledges grant RYC2022-036010-I funded by MICIU/AEI/10.13039/501100011033 and ESF+.
 
\section*{Ethics statement}
 
This study uses public anonymized data from The Cancer Imaging Archive \citep{noauthor_acrin_nodate}. No identifiable patient information was used in this work.
 
\section*{Declaration of competing interest}
 
The authors declare that they have no known competing financial interests or personal relationships that could have appeared to influence the work reported in this paper.
 
\section*{Data availability}
 
The imaging and clinical data used in this study are publicly available from The Cancer Imaging Archive \citep{noauthor_acrin_nodate}. The post-processed DCE-MRI tumor masks are publicly available from Zenodo \citep{corral_fontecha_2026}.

\section*{Declaration of generative AI and AI-assisted technologies in the manuscript preparation process}

During the preparation of this work, the authors used OpenAI Codex solely to review formal aspects of the manuscript, including language, internal consistency, formatting, and compliance with journal guidelines. After using this tool, the authors reviewed and edited all suggestions as needed and take full responsibility for the content of the publication.
 
\renewcommand{\bibfont}{\footnotesize}
\balance
\bibliography{bibliography}
 
\end{document}